\documentclass{article}
\usepackage{amsmath,epsfig}
\usepackage[preprint]{spconfa4}
\usepackage{atbegshi}

\AtBeginDocument{\AtBeginShipoutNext{\AtBeginShipoutDiscard}}

\copyrightnotice{\textbf{978-1-7281-1331-9/20/\$31.00~\copyright 2020 IEEE}}

\let\OLDthebibliography\thebibliography
\renewcommand\thebibliography[1]{
  \OLDthebibliography{#1}
  \setlength{\parskip}{0pt}
  \setlength{\itemsep}{0pt plus 0.3ex}
}

\begin{document}\sloppy

\def\x{{\mathbf x}}
\def\L{{\cal L}}
\def\etal{\emph{et. al.}}

\title{Region-Adaptive Texture Enhancement for Detailed Person Image Synthesis}
%
\name{Lingbo Yang$^1$, Pan Wang$^2$, Xinfeng Zhang$^3$, Shanshe Wang$^1$, Zhanning Gao$^2$}\nametwo{Peiran Ren$^2$, Xuansong Xie$^2$, Siwei Ma$^1$, Wen Gao$^1$}
\address{$^1$ Institute of Digital Media, School of Electronic Engineering and Computer Science, Peking University\\ $^2$ Artificial Intelligence Center, DAMO Academy, Alibaba Group \\$^3$ School of Computer Science and Technology, University of Chinese Academy of Sciences}

\thanks{This work was done when Lingbo Yang is working at DAMO Academy,
Alibaba as a research intern. Siwei Ma is the corresponding author. This
work was supported by the National Natural Science Foundation of China
(61632001) and High-performance Computing Platform of Peking University,
which are gratefully acknowledged.}

\maketitle

\begin{abstract}
    The ability to produce convincing textural details is essential for the fidelity of synthesized person images.
Existing methods typically follow a ``warping-based'' strategy that propagates appearance features through the same pathway used for pose transfer. However, most fine-grained features would be lost due to down-sampling, leading to over-smoothed clothes and missing details in the output images.
In this paper we presents RATE-Net, a novel framework for synthesizing person images with sharp texture details. The proposed framework leverages an additional texture enhancing module to extract appearance information from the source image and estimate a fine-grained residual texture map, which helps to refine the coarse estimation from the pose transfer module. In addition, we design an effective alternate updating strategy to promote mutual guidance between two modules for better shape and appearance consistency. Experiments conducted on DeepFashion benchmark dataset have demonstrated the superiority of our framework compared with existing networks. 

\end{abstract}

\section{Introduction}

In this paper, we study the problem of synthesizing images of a person's appearance under novel poses, which is commonly known as \emph{human pose transfer}~\cite{PG2,DeformableGAN,PATN}.
The problem is first introduced in~\cite{PG2}, where a transfer model receives a source image that provides conditional appearance constraints, and is expected to transfer the person's appearance to new poses. Human pose transfer forms the core of many real-world applications, including interactive fashion design, creative media production, and many other human-centered tasks.

In a typical person image, the bulk of the area is occupied by clothes with rich textural patterns, which are more likely to attract a viewer's attention. Therefore, the ability to synthesize realistic texture details is crucial for the performance and human perception of a transfer model. In the area of image-to-image~(I2I) translation~\cite{pix2pix,cyclegan}, the U-net architecture~\cite{unet} has achieved remarkable success in preserving fine-grained visual details through skip-connections. For human pose transfer task, however, such technique is not suitable due to the structural deformation of the human body under different poses. Siarohin~\etal proposed a modified version called deformable skip-connection~\cite{DeformableGAN}, where feature maps extracted from the source image are warped onto the corresponding target regions according to pose correspondences.

\begin{figure}[t]
  \centering
  \includegraphics[width=\linewidth]{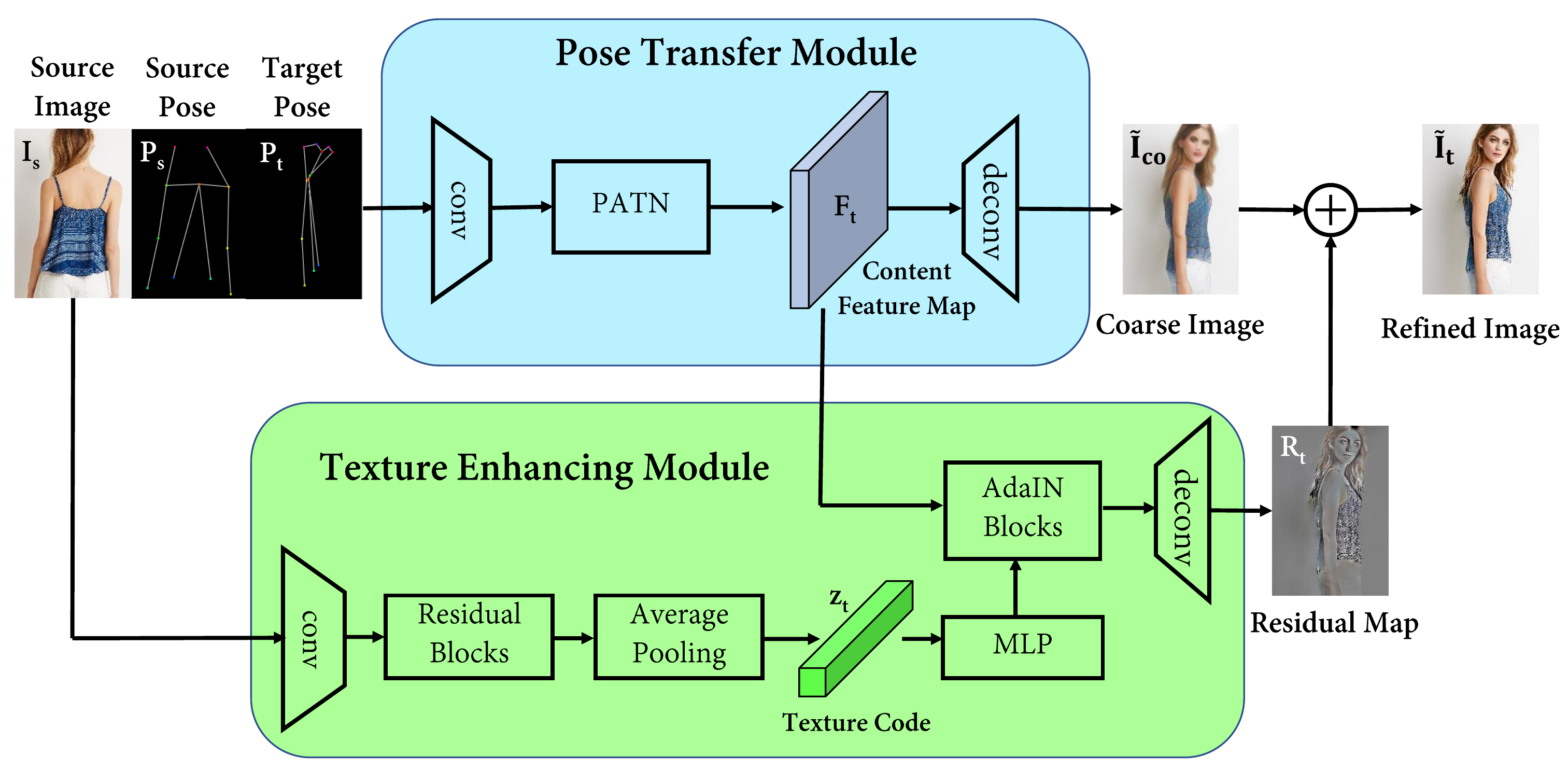}
  \caption{Overview of our proposed RATE-Net. The pose transfer module first estimates a coarse output $\tilde{I}_{co}$ under the target pose. In addition, it also provides regional guidance for the texture enhancing module, which then synthesizes a residual map $R_t$ to refine the coarse output.}\label{fig:framework}
\end{figure}

The deformable skip-connections~\cite{DeformableGAN} encapsulates the two most distinctive features of the so-called ``warping-based'' methods:
(1) A pixel-wise (part-wise) correspondence map estimated from the source pose to the target pose;
(2) A mechanism for warping features from the source image to corresponding target regions according to the estimated mapping.
However, most existing methods~\cite{DensePoseTransfer,softgate,DIAF,liquidwarpinggan} propose to channel the appearance features through the same pathway used for pose transfer, where the loss of fine-grained textural information would be inevitable due to down-sampling operations. This often leads to over-smoothed clothes and distorted facial landmarks that severely degrade the visual quality of synthesized images. In addition, estimating pixelwise mapping is extremely time-consuming and often requires additional dense annotations, making it intractable for many interactive designing and editing applications.

Instead, we seek another path of enhancing appearance details of pose-transferred images by synthesizing new textures that match the style of the input source image. To this end, we propose a novel \emph{Region-Adaptive Texture Enhancing Network} (RATE-Net), where an additional texture enhancing module is introduced to estimate a residual map for detail refinement upon coarse estimations by the pose transfer module. The architecture of our framework is shown in Fig~\ref{fig:framework}, where the source image is utilized in two ways: For the pose transfer module, we estimate a feature map roughly aligned with target pose, which later serves as the spatial guidance for the texture enhancing module. In addition, we also extract the style and textural information from the source image into compact codes, and inject the codes into the residual enhancing map through adaptive normalization layers~\cite{adain}. To help strengthen the mutual guidance between two modules, we also design an alternate training strategy to further improve the overall performance. Extensive experiments conducted on the challenging DeepFashion~\cite{DeepFashion} benchmark dataset demonstrate the superiority of the proposed framework against recent warping-based methods. In summary, our contributions are twofold:

\begin{itemize}
  \item We propose RATE-Net, a novel enhancing based solution that utilize the style and texture information from the input source image for better textural refinement upon coarse images. The network utilizes the source image for both label map estimation and texture/style control, which leads to better pose transfer results compared with warping-based methods.
  \item We design an effective training strategy to maximize the mutual guidance between two modules, where pose transformation mapping can be further refined through the style-aware loss function computed over enhanced images, thus helps to preserve the integrity of human body.
\end{itemize}

\section{Related Works}

\textbf{Human Pose Transfer} is first described in PG2~\cite{PG2} where the goal is to transfer the appearance of a person from the source image into new poses.
Existing works typically focus on warping the textures from the source image to the target image based on the warping transformation estimated between the corresponding poses. DSC~\cite{DeformableGAN} segments the human skeleton into rigid parts and approximates the warping function with piece-wise affine transformation. Some recent works~\cite{DensePoseTransfer,softgate,DIAF,liquidwarpinggan} further estimate pixel-level feature warping flow by leveraging dense keypoint annotations, such as DensePose~\cite{densepose} and SMPL~\cite{SMPL}.
However, estimating dense annotations and pixel-wise warping flow are typically computationally expensive, and the corresponding groundtruth annotations are more difficult to collect. PATN\cite{PATN} proposed a lightweight network to gradually transfer a person's pose through several cascaded pose-attentional transfer blocks. However, most appearance details of the input image would be lost due to the down-sampling operation, which often leads to inferior results when dealing with person images with texture-rich clothes. Instead, we propose a region adaptive texture enhancing network which is better at capturing and re-synthesizing fine-grained details. Furthermore, it only relies on 2D skeletons for pose annotation and can be trained in an end-to-end fashion.

\textbf{Adaptive Normalization} provides a new mechanism to inject guidance information into the main image generation pathway. It is typically implemented as an affine transformation over normalized feature responses, with parameters inferred from external data. AdaIN~\cite{adain} was first proposed in style transfer task, and later used in other tasks such as high resolution face image synthesis~\cite{stylegan} and few-shot I2I translation~\cite{funit}. SPADE~\cite{spade} further expands the mean and deviation parameters from vectors to 3-tensors, thus incorporating spatial attention into semantic controlled image synthesis. This idea is also useful for few-shot video synthesis~\cite{fewshotvid2vid}. Inspired by these works, we incorporate the pose guidance into our appearance encoding network to capture fine-grained visual information from texture-rich regions for detail enhancement in synthesized images.

\section{The Network Architecture}

The proposed RATE-Net contains two modules: A pose transfer module that generates a coarse image under the target pose, and a texture enhancing module that estimates a residual map to fill in more appearance details onto the coarse image. The overall architecture is illustrated in Fig.~\ref{fig:framework}.

\subsection{Notations}
We first introduce some notations. The network takes two inputs, a source image $I_s$ and a target pose $P_t$, and tries to generate a new image $I_t$ containing the person in $I_s$ under the pose $P_t$. An $18\times 2$ array of keypoint coordinates is estimated for both the source and target image, denoted as $P_s$ and $P_t$, respectively. During the training, the generator $G$ is fed with paired images $(I_s, I_t)$ of the same person under different poses along with the corresponding pose heatmaps $(P_s, P_t)$, which can be estimated and cached before training. The output of the network, $\tilde{I_t} = G(I_s; P_s, P_t)$, is compared with groundtruth image $I_t$ for losses. Below we further dissect the generator $G$ and elaborate on the two modules.

\subsection{Pose Transfer Module}\label{section:pose}
As shown in Fig.~\ref{fig:framework}, the input of the pose transfer module consists of a source image $I_s$ and a pair of target poses~$(P_s, P_t)$, which are concatenated along the channel dimension. The network consists of several convolutional down-sampling layers, followed by a series of pose-attentional transfer blocks~\cite{PATN} that help to warp the contents of the source image onto the target pose $P_t$ in a progressive manner. The output feature map $F_t$ is then fed into the up-sampling layers to recover a coarse estimation $\tilde{I}_{co}= G_P(I_s; P_s, P_t)$ of the target image $I_t$.

It should be emphasized that although our pose-transfer module shares a similar network architecture as in~\cite{PATN}, the purpose and training strategy are completely different. In particular, we DO NOT aim to directly \emph{warp} the fine-grained textures from the source image into the target pose, as most of the textural information would be lost due to the down-sampling operation. Instead, we utilize the pose-transfer network to acquire a reasonable content feature map $F_t$ under the target pose, which provides regional guidance for the preceding texture enhancing module by hinting \emph{``where to add more texture details''}. In this way, our framework can \emph{synthesize} new textures with a separate network and build the final pose-transferred image in a \emph{coarse-to-fine} manner, which greatly improves the training stability and the performance of our network. Ablation studies are presented to validate our claims.

\subsection{Texture Enhancing Module}
The texture enhancing module aims to recover fine-grained visual details from the source image $I_s$, and enhance the coarse estimation $\tilde{I}_{co}$ by synthesizing a region-aware residual texture map $R_t = G_t(I_s; F_t)$ under the guidance of pose-aligned content feature map $F_t$. We reuse the source image $I_s$ to extract texture codes $z_t$ with an encoder containing several convolutional down-sampling layers, residual blocks and an average pooling operation. To inject the textural code into the content feature map, we utilize the Adaptive Instance Normalization~\cite{adain} that originates from style transfer tasks. In particular, $z_t$ now plays the role of ``style guide'' which controls the pattern and granularity of textures in respective regions. The final output is acquired by adding the residual map $R$ onto the coarse estimation:
$$\tilde{I}_t = \tilde{I}_{co} + R_t = G_P(I_s; P_s, P_t) + G_t(I_s; F_t)$$

\subsection{Discriminators}
We leverage the design in~\cite{PATN} that uses two discriminators to differentiate real and fake samples both in terms of shape and appearance consistency. The shape discriminator $D_S$ evaluates input image/pose pairs for shape consistency $R_S = D_S(\tilde{I}_t; P_t)$, and the appearance discriminator $D_A$ compares the appearance consistency between the synthesized image and the source image $R_A = D_A(\tilde{I}_t; I_S)$. Unlike~\cite{PATN} that multiplies the scores, we train the two discriminators separately so that both criterion can be individually analyzed and optimized.

\section{The Training Strategy}

It is clear from Fig.~\ref{fig:framework} that the two modules of our proposed framework are mutually dependent: The texture enhancement module relies on the pose-aligned guidance map $F_t$ to put textures on the right places, and the losses computed over texture-enhanced images are back-propagated to the pose transfer module, which helps improve the accuracy of the estimated guidance map. Therefore, we figure that using an alternate updating strategy should be useful for promoting the mutual guidance between two modules and achieve the best overall performance. Concretely, for each input batch, we first update the pose transfer module with a loss function $\L_1$ defined over coarse estimation $\tilde{I}_{co}$ before performing an end-to-end fine-tuning step to update two modules together with another texture-aware loss $\L_2$ defined over final output $\tilde{I}_t$. The discriminator is then updated for $K$ steps, where we empirically found $K=3$ leads to a nice balance between training speed and discriminative capability. This concludes a ``1-1-3'' training cycle of our framework.

\subsection{Loss Functions}
As discussed in section~\ref{section:pose}, the pose transfer module is not designed for conveying fine-grained texture information, and the estimated coarse output can be expected to lack visual realism. Therefore it's unnecessary to enforce discriminative loss in $\L_1$. Instead, we simply adopt the following formulation:
$$\L_1 = \lambda_{recon}\L_{recon} + \lambda_{per}\L_{per}$$
where $\L_{recon}$ is the pixelwise L1 loss, and $\L_{per}$ is the perceptual loss in~\cite{PerceptualLoss}:

$$\L_{per} = \frac{1}{CHW}\sum_{l}w_l\|\phi_l(I_t) - \phi_l(\tilde{I}_{co})\|_1$$
Here $\phi$ is a pretrained VGG-19 network and $l$ denotes the layer index. In practice we found it effective to sample from different layers and use the weighted average loss to balance the perceptual consistency across different scales.

For loss function in step 2, we further add style loss and adversarial loss terms upon $\L_1$, leading to the full loss function as follows:

$$\L_2 = \lambda_{recon}\L_{recon} + \lambda_{per}\L_{per} + \lambda_{sty}\L_{sty} + \lambda_{GAN}\L_{GAN} $$
where $\L_{sty}$ is the Gram-matrix based style loss~\cite{PerceptualLoss} and $\L_{GAN}$ is the addition of losses in $D_S$ and $D_A$, which is also used for discriminator updating. Readers can find more details in the corresponding literature.

\subsection{Implementation Details}
We implement the proposed framework in PyTorch. Both modules include 3 down-sampling layers, and the pose transfer module contains 9 cascaded pose transfer blocks. LeakyReLU is used after convolution and normalization layers with 0.2 negative slope. The dimension of the texture codes is set to 128, and the impact of different lengths are further analyzed in ablation study. We use the Rectified Adam optimizer~\cite{radam} for better stability and performance at convergence. The full training involves 40K alternating cycles leading to a total of 200K iterations. The learning rate is set to $1e-4$ for all networks, and is fixed for the first 10K cycles before linearly drops to 0. The weight $\lambda_{recon}$ is set to 10 and other weights are set to 5.

\begin{figure}[t]
  \centering
  \includegraphics[width=\linewidth]{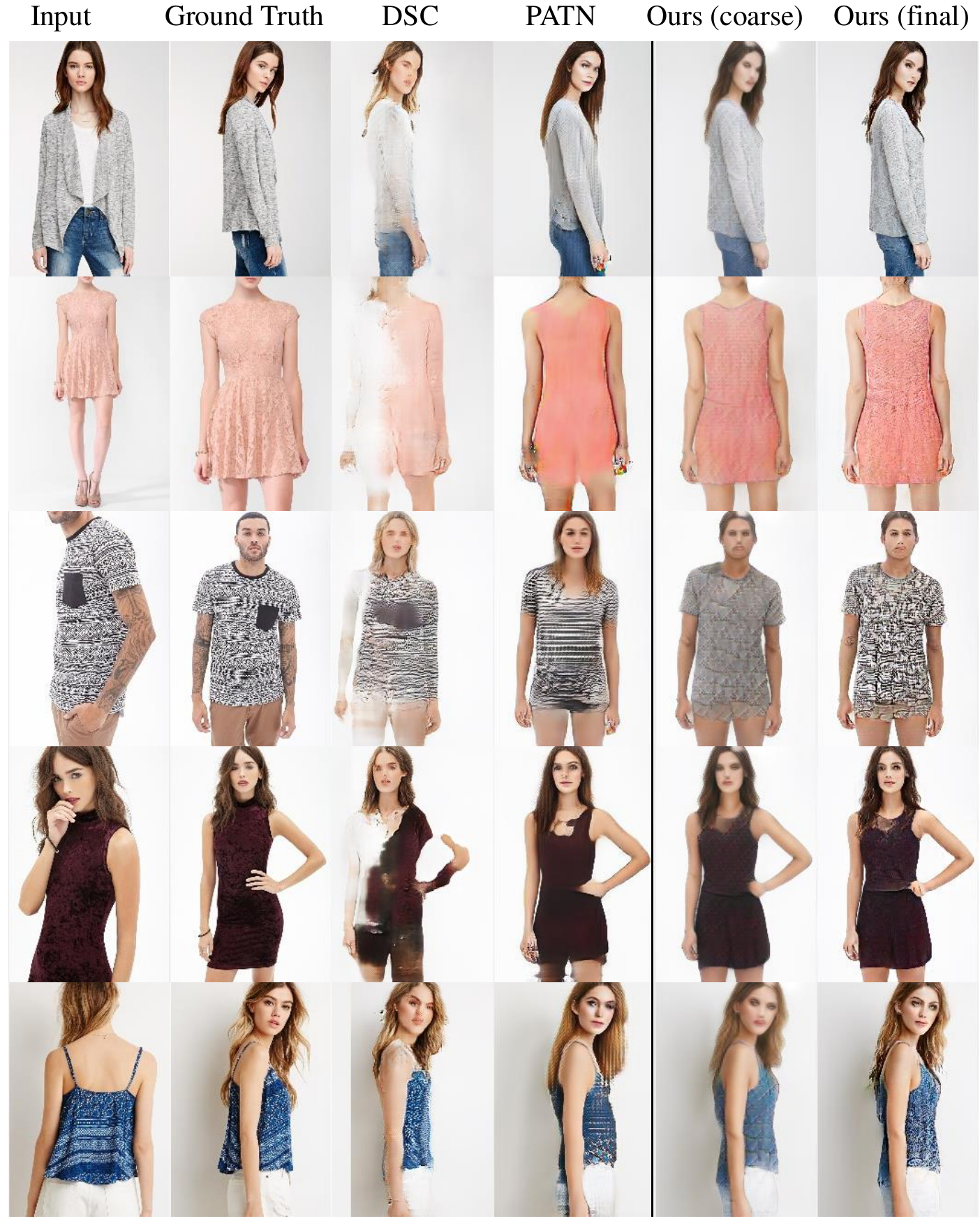}
  \caption{Qualitative results of the proposed method against several competitive baselines. Some images have been cropped for visualization purposes. \emph{Please zoom in for details.}}\label{fig:sota}
\end{figure}

\section{Experiments}

In this section, we compare our framework with several state-of-the-art methods to demonstrate the superiority of our framework. Furthermore, we perform a thorough ablation study to verify the efficacy of our main contributions.

    \textbf{Dataset} We validate our proposed framework on the \emph{In-shop Clothes Retrieval Benchmark} of DeepFashion~\cite{DeepFashion} containing about 50K images of fashion models in texture-rich clothes under various poses. The images are in 256 $\times$ 256 resolution and contain clean background. We adopt Openpose~\cite{Openpose} to estimate an 18-point skeleton for each image, and convert it into a 18-channel heatmap as in ~\cite{DeformableGAN}. We use the train/test pairs in~\cite{PATN}, where 101,966 pairs are randomly selected for training and 8,570 pairs for testing. The partition guarantees that the persons appeared in training and testing sets do not overlap, making it more reliable to validate the generalization ability of our model.

    \textbf{Evaluation Metrics} For human pose transfer task, we aim to evaluate both the statistical fidelity and the perceptual quality of generated images. To this end, we adopt the Structural Similarity (SSIM)~\cite{SSIM} and Inception Score (IS)~\cite{InceptionScore} to account for the model's performance in both perspectives. However, a recent study~\cite{ISnote} has pointed out that IS is theoretically flawed and cannot always provide useful guidance when comparing models. To more reliably evaluate the perceptual quality of generated images, we introduce two more \emph{supervised} metrics: FID~\cite{FID} and LPIPS~\cite{LPIPS}. Note that both metrics utilize a pretrained network to convert the images into feature space, and compute the distance between image features with respect to both the global distribution and each pair of samples. We believe that \emph{supervised} perceptual metrics can better reflect the perceptual fidelity of our proposed model than the \emph{unsupervised} IS metric.

\begin{table}[t]
\centering
\begin{tabular}{c|cccc}
\hline
Models & SSIM $ \uparrow $ & IS $ \uparrow $ & FID $ \downarrow $ & LPIPS $ \downarrow $ \\
\hline
UV-Net    & 0.763 & \underbar{3.440} & --- & --- \\
SPT       & 0.736 & \textbf{3.441} & --- & --- \\
DSC   & 0.718 & 2.978 & 54.789 & 0.496 \\
PATN        & 0.773 & 3.209 & 19.816 & 0.253 \\
\hline
Ours Coarse        & \textbf{0.780} & 3.230 & \underbar{18.405} & \underbar{0.243}\\
Ours Full        & \underbar{0.774} & 3.125 & \textbf{14.611} & \textbf{0.218}\\
\hline
\end{tabular}
\caption{Performance against other baseline methods, the highest and the second highest value for each metric is shown in bold / underline format. Up arrow means higher score is preferred, and vice versa.}
\label{tab:compare_sota}
\end{table}

\begin{table}[t]
\centering
\begin{tabular}{c|cccc}
\hline
Models & SSIM $ \uparrow $ & IS $ \uparrow $ & FID $ \downarrow $ & LPIPS $ \downarrow $ \\
\hline
PB Only          & 0.772 & \textbf{3.231} & 18.602 & 0.250 \\
PB Fixed         & 0.767 & 2.923 & 20.346 & 0.238 \\
Texture64dim     & 0.763 & 3.040 & 18.263 & 0.243 \\
Ours Full        & \textbf{0.774} & 3.125 & \textbf{14.611} & \textbf{0.218}\\
\hline
\end{tabular}
\caption{Quantitative ablation study results, the highest value for each metric is shown in bold format.}
\label{tab:ablation}
\end{table}

\subsection{Comparison with Previous Works}

 We compare our proposed framework with several representative works: DSC~\cite{DeformableGAN}, UV-Net~\cite{UVNet}, SPT~\cite{SPT} and PATN~\cite{PATN}. All the tests are carried out on the same set of testing pairs in~\cite{PATN}.
 Table \ref{tab:compare_sota} shows significant improvement of the proposed framework against recent state-of-the-art methods in terms of perceptual quality, while the SSIM and IS scores are also comparable.
 To further analyze the impact of the texture enhancing module, we also evaluate the quality of coarse estimations without residual map. As observed in Table~\ref{tab:compare_sota}, although the SSIM score slightly drops after enhancement, the perceptual fidelity is significantly improved, with 20\% gain on FID and 10\% gain on LPIPS. Furthermore, although the network backbone is highly similar, our pose transfer module still performs considerably better than the original PATN implementation in perceptual fidelity, which can verify the efficacy of our texture enhancing module for refining the estimation of the guidance map $F_t$.

 In addition, we also showcase the qualitative result for some challenging examples with large pose transition and texture-rich garments, As shown in Fig~\ref{fig:sota}, our method is more faithful to the input pose and appearance condition than other baselines and possesses more realistic visual details, especially like clothing textures and hair waves. Also, the gender bias issue~\cite{DeformableGAN} on DeepFashion dataset is partly resolved as shown in the third row, where all the other methods output a female image by mistake. Moreover, it can be observed that the texture enhancing module is effective for recovering fine-grained visual details lost in pose transfer module and consistently improving the visual quality of synthesized human images.

\begin{figure}[t]
  \centering
  \includegraphics[width=\linewidth]{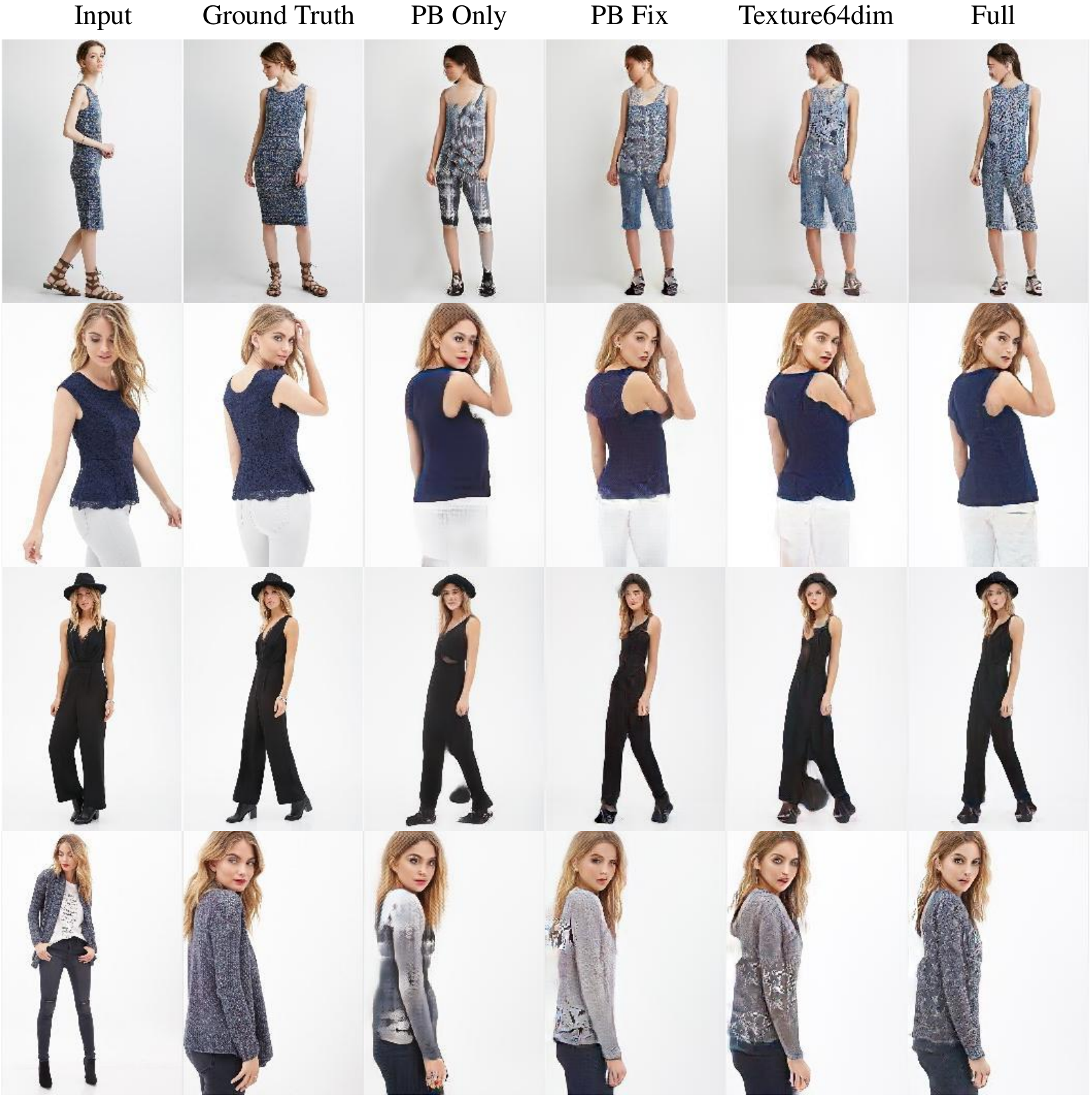}
  \caption{Qualitative results of ablation methods. \emph{Please zoom in for details.}}\label{fig:ablation}
\end{figure}

\subsection{Ablation Study}
In this section, we perform an ablation study to analyze the effect of different parts of our proposed framework on the final performance. For each part, we set up a corresponding baseline by either removing this part from the whole framework or changing the key parameters.

\textbf{PB Only} We remove the texture enhancing module and train the pose-transfer module directly in an end-to-end fashion with loss function $\L_2$ computed over coarse estimation $\tilde{I}_{co}$. Notice that this setting slightly differs from the PATN~\cite{PATN} baseline as the loss function has an additional style loss term, and the weights are slightly modified.

\textbf{PB Fix} To further investigate the efficacy of our proposed training scheme, we initialize the parameters of the pose-transfer module using pretrained models in~\cite{PATN}, and keep it fixed during training. In this way, the interaction between two modules is broken, and detailed style loss and conditional adversarial losses cannot be back-propagated into the pose transfer module.

\textbf{Texture64dim} We construct the corresponding baseline by reducing the length of the textural code to 64. In this way, the textural information extracted from source image is reduced, which could result in less informative textures and more severe artifacts.

\textbf{Full} This is the full version of our proposed framework used for comparing with other SOTA methods.

We report the quantitative performances of all four methods on the DeepFashion dataset in Table~\ref{tab:ablation}. As observed, our full method has a clear advantage over other ablation methods, especially in terms of perceptual distance metrics like FID and LPIPS, which we believe has highlighted the importance of textural enhancement in human pose transfer task. It is also noteworthy that increasing the length of texture code can significantly boost the performance, and we speculate that this could lead to a useful method for determining the \emph{intrinsic dimension} of the latent texture space, which is the length of the texture code at which point the model's performance stops improving. We didn't evaluate our model at higher code lengths because the memory cost had become intolerable. However, we believe that it's still possible to further improve the performance by using longer texture codes.

To better visualize the impact of different components, we showcase some representative examples in Fig.~\ref{fig:ablation}. As observed, our full framework is capable of synthesizing much finer textural details than all other methods. In addition, our framework also better preserves the integrity of human body (as observed in the second and the third row), which to our belief indicates the increasing accuracy of the pose-aligned feature map $F_t$ due to more localized guidance provided by the loss function $\L_2$ defined over texture enhanced images. This further justifies our claim of mutual guidance between two modules and the proposed alternate training strategy.

\section{Conclusion}
We presents RATE-Net, a novel framework for synthesizing person images with sharp texture details. 
Instead of simply warping the patches from the source image with the risk of losing fine-grained texture details, we proposed to synthesize new textures with additional texture enhancing module that helps add more visual details to the coarse pose transfer results. Furthermore, an effective training strategy is proposed to alternately update the two modules for better overall performance. Compared with previous works, our framework can synthesize human images with much finer details, and can better preserve the style and appearance of the source image.
We believe the idea behind our framework can inspire other related topics in semantic image synthesis as well.


\bibliographystyle{IEEEbib}
\bibliography{icme2020blind}

\end{document}